\pdfoutput=1

\documentclass[11pt]{article}
\usepackage[final]{acl}

\usepackage{times}
\usepackage{latexsym}
\usepackage{multirow} 
\usepackage{booktabs}

\usepackage[T1]{fontenc}

\usepackage[utf8]{inputenc}

\usepackage{microtype}

\usepackage{inconsolata}

\usepackage{graphicx}

\usepackage{enumitem}

\usepackage{tabularx}
\usepackage{amsmath}

\usepackage{hyperref}
\usepackage{bbm}
\usepackage{linguex}

\usepackage[table]{xcolor}
\usepackage{booktabs}

\definecolor{comp5}{RGB}{193,0,13}
\definecolor{comp4}{RGB}{255,136,41}
\definecolor{comp3}{RGB}{255,214,107}
\definecolor{comp2}{RGB}{165,201,15}
\definecolor{comp1}{RGB}{111,156,61}

\title{Are BabyLMs Deaf to Gricean Maxims? \\ A Pragmatic Evaluation of Sample-efficient Language Models}

\author{
    \textbf{Raha Askari$^{1,2}$,} 
    \textbf{Sina Zarrieß$^2$,}
    \textbf{Özge Alacam$^2$,}
    \textbf{Judith Sieker$^2$}\\
    $^1$Department of Humanities, University of Turin\\
    $^2$Computational Linguistics, Department of Linguistics, Bielefeld University
    }

\begin{document}

\maketitle

\begin{abstract}
Implicit meanings are integral to human communication, making it essential for language models to be capable of identifying and interpreting them. \citet{grice1975logic} proposed a set of conversational maxims that guide cooperative dialogue, noting that speakers may deliberately violate these principles to express meanings beyond literal words, and that listeners, in turn, recognize such violations to draw pragmatic inferences.
Building on \citet{article}’s study of children’s sensitivity to violations of Gricean maxims, we introduce a novel benchmark to test whether language models pretrained on <10M and <100M tokens 
can distinguish maxim-adhering from maxim-violating utterances. We compare these BabyLMs across five maxims and situate their performance relative to children and a Large Language Model (LLM) pretrained on 3T tokens.
We find that overall, models trained on <100M tokens outperform those trained on <10M, yet fall short of child-level and LLM competence. 
Our results suggest that modest data increases improve some aspects of pragmatic behavior, leading to finer-grained differentiation between pragmatic dimensions.

Our benchmark extends the BabyLM evaluation suite to pragmatic aspects of language and is publicly available.\footnote{\url{https://huggingface.co/datasets/rahaaskari/gricean_baby}}

\end{abstract}

\section{Introduction}

Consider the following exchange:
Sarah asks her friend \textit{"What did you eat for lunch?"}, upon which her friend might reply \textit{"I had something edible"} or \textit{"I had chicken soup with an extra small silver spoon"}.
While both responses are true and perfectly grammatical, the first one fails to provide the amount of information Sarah's question calls for, and the second one contains excessive, unasked details. Most listeners would expect an answer that is specific but not unnecessarily detailed, such as \textit{"I had chicken soup"}.

In everyday conversation, such under- or over-informative replies stand out as odd because they do not provide the adequate amount of detail the question asks for.
The philosopher \citet{grice1975logic} explained such phenomena through his \textit{Cooperative principle}, which holds that speakers are generally aware of what is conversationally suitable or unsuitable. He proposed a set of conversational maxims, one of which, the \textit{maxim of Quantity}, requires the speaker to be as informative as necessary.  

The ability to notice and interpret such deviations from conversational norms is a key aspect of pragmatic competence, and essential for successful communication.
In the evaluation of Large Language Models (LLMs), however, while models are now routinely tested on a wide range of syntactic tasks (e.g., \citet{marvin-linzen-2018-targeted, hu-etal-2020-systematic, finlayson-etal-2021-causal, lampinen-2024-language,  kryvosheieva-levy-2025-controlled}), far fewer studies target their ability to reason pragmatically \cite{Ettinger2020, fried-etal-2023-pragmatics, ma-etal-2025-pragmatics, Sieker2025, lachenmaier-etal-2025-llms}.
This gap is especially pronounced for resource-limited models such as those developed for the BabyLM Challenge \citep{warstadt2023papersbabylmchallenge, choshen2024callpapers2ndbabylm}. 
One possible reason is that, unlike syntax, pragmatics does not easily lend itself to large-scale minimal-pair test creation. Controlled operationalizations of pragmatic phenomena, such as those in \citet{sieker-etal-2023-beyond} and \citet{sieker-zarriess-2023-antipsps}, remain rare and resource-intensive, highlighting the challenge of designing systematic evaluation materials for this domain.

In psycholinguistics, however, several diagnostic tasks for pragmatic understanding already exist (e.g., \citet{doran2012novel, degen, Romoli2015, Tieu2015}). 
One such task is the Conversational Violations Test (CVT), designed to investigate children's pragmatic abilities based on Gricean maxims, introduced in \citet{article}'s study \textit{"Are Children with Autism Deaf to Gricean Maxims?"}.
In the CVT, children are presented with short dialogues where one answer follows one of Gricean conversational maxims and another violates it. Children are asked to identify the maxim-violating response. 
This controlled forced-choice format makes the CVT particularly attractive for LM evaluation: the correct choice depends on recognizing conversational norms rather than relying solely on factual knowledge or grammar, and the task fits well to established evaluation methods that compare model-assigned probabilities for a predefined set of candidate responses, as in grammatical acceptability \citep{warstadt-etal-2020-blimp-benchmark}, abductive commonsense reasoning \citep{zhao-etal-2023-abductive}, or semantic relations of compound nouns \citep{rambelli-etal-2024-large}.

In this paper, we adapt the CVT into a benchmark for evaluating BabyLMs' sensitivity to Gricean maxims. Starting from the original 25 conversational items from CVT, we augment the dataset automatically to over 2,250 items and refine them through human annotation.
We evaluate a range of BabyLM baseline models (4 trained on <10M and 4 trained on <100M words), compare their performance to that of children from \citet{article}, and situate their results alongside an LLM trained on more extensive data.  
In total, our experiment produces 20,250 data points (across 8 BabyLMs and 1 LLM).
Among the evaluated models, BabyLMs trained on <100M tokens outperformed those trained on <10M, yet both groups fell short of achieving child-level pragmatic accuracy despite their developmental motivation. On average, BabyLMs performed best when judging truthfulness, but struggled most with assessing the appropriate level of informativeness. The LLM generally surpassed the BabyLMs and, in some cases, even outperformed children, but still failed to match children’s overall competence. Thus, despite vastly larger training data, notable gaps persist between model and child performance across several maxims.

The contributions of this study are threefold: (1) a novel, linguistically controlled benchmark for testing pragmatic competence in data-limited LMs, grounded in established psycholinguistic methodology; (2) an empirical analysis of BabyLMs’ performance across different Gricean maxims; and (3) a comparison of LM and child performance that situates model behavior within a developmental trajectory. 

\section{Background}

Effective communication relies on more than just producing grammatical sentences. 
Much of what speakers communicate is conveyed implicitly, relying on the listener to infer meanings that go beyond the literal words. To do so successfully, speakers must choose utterances that are appropriate to the conversational context, and listeners must interpret them in light of shared assumptions, intentions and social norms. Even a perfectly well-formed sentence can be unhelpful, misleading or socially awkward if it ignores these unspoken rules.
The study of how meaning is shaped by such contextual factors is the domain of pragmatics, which rests on the central idea that conversation is a cooperative activity: participants work together to exchange information efficiently and meaningfully.
\citet{grice1975logic} formalized this intuition in his \textit{Cooperative Principle}, according to which interlocutors are generally aware of what is conversationally suitable or unsuitable at each stage of a dialogue. 
He categorized this principle into four maxims, and additionally discussed Politeness as what he termed an "off-the-list" maxim:

\begin{itemize}
    \item Quantity (\textit{Make your contribution as informative as required and do not make your contribution more informative than is required})
    \item Quality (\textit{Do not say what you believe to be false and do not say that for which you lack adequate evidence})
    \item Relation (\textit{Be relevant})
    \item Manner (\textit{Be perspicuous, i.e., avoid obscurity, avoid ambiguity, be brief and be orderly}).
    \item Politeness (\textit{Be polite})
\end{itemize}

While these maxims are typically adhered to, speakers may sometimes blatantly violate them by saying one thing but implying another, producing what is known as an \textit{implicature}.
For example, when two colleagues are talking during a lunch break, one might ask \textit{"Did you talk to the boss about the promotion?"}, and the other might reply, \textit{"I really like this food."} 
This response violates the maxim of Relation and prompts the listener to search for the intended meaning, assuming the other person remains cooperative and aware of the maxims. In this case, for example, the interlocutor is likely to infer that their colleague wishes to avoid the topic and has not spoken to the boss.
Over the past decades, the Gricean maxims have become a cornerstone of pragmatic theory, shaping how researchers analyze and explain the ways people interpret and produce language in context.

\paragraph{Developmental Studies.}
Several developmental psycholinguistic studies have examined the age at which the sensitivity to such conversational violations emerges in humans (to name a few;  \citealp{ackerman1981question}; \citealp{conti1984children}; \citealp{article}; \citealp{surian2010sensitivity}; \citealp{okanda2015understanding} and \citealp{panzeri2021children}). In this direction, \citet{article}'s study introduced the Conversational Violations Test (CVT) to compare the pragmatic abilities of children with autism and specific language impairments to those of neurotypical children by incorporating Grice's framework. 
The maxims addressed in their study were Quantity (divided into two maxims; I: \textit{Be informative} and II: \textit{Avoid redundant information}), Quality, Relation and Politeness. The CVT is a set of 25 short conversational items and contains 5 conversations for each maxim.
In their experiment, 8 neurotypical children (mean age 6-7) were presented with tape-recorded conversations featuring three puppets. In each scenario, one puppet would ask a question, and the other two would respond, only that one of them would provide an answer that violated a conversational maxim. The children were then asked to identify the puppet that gave the \textit{silly or funny} answer, i.e., the one that violated the maxim. See Table \ref{tab:cvt_examples} for examples for each maxim.

\begin{table}[t]
\centering
\begin{tabular}{p{0.23\linewidth} p{0.64\linewidth}}
\toprule
\textbf{Maxim} & \textbf{Example} \\

\midrule
\textbf{Quantity\,I}\newline \textit{Be informative} &
\textbf{Q:} \textit{How do you prefer your tea?} \newline
\textbf{Follower:} \textit{With milk.} \newline
\textbf{Violator:} \textit{In a cup.} \\

\midrule
\textbf{Quantity\,II}\newline \textit{Avoid redundant information}&
\textbf{Q:} \textit{Who is your best friend?} \newline
\textbf{Follower:} \textit{My best friend is John. He goes to my school.} \newline
\textbf{Violator:} \textit{My best friend is Peter. He wears clothes.} \\

\midrule
\textbf{Quality}\newline \textit{Be truthful} &
\textbf{Q:} \textit{Where do you live?} \newline
\textbf{Follower:} \textit{I live in London.} \newline
\textbf{Violator:} \textit{I live on the moon.} \\

\midrule
\textbf{Relation}\newline \textit{Be relevant}&
\textbf{Q:} \textit{What games do you know?} \newline
\textbf{Follower:} \textit{I know how to play football.} \newline
\textbf{Violator:} \textit{I know your name.} \\

\midrule
\textbf{Politeness}\newline \textit{Be polite}&
\textbf{Q:} \textit{Do you like my dress?} \newline
\textbf{Follower:} \textit{It's pretty.} \newline
\textbf{Violator:} \textit{I hate it.} \\

\bottomrule
\end{tabular}
\caption{Example items for different conversational maxims from \citet{article}'s CVT. The \textit{Follower} adheres to the maxim, while the \textit{Violator} does not.}
\label{tab:cvt_examples}
\end{table}

\paragraph{BabyLMs.} 
The BabyLM Challenge 
aims to model human language development in order to optimize language model pretraining under strict data limitations. Tracks for submissions of text-only models include the Strict-small track (trained on <10M tokens) and the Strict track (trained on <100M tokens). As a starting point for evaluation, the organizers release a group of baseline models accompanied with pretraining corpora, along with an evaluation pipeline including benchmarks such as BLiMP \citep{warstadt-etal-2020-blimp-benchmark} or GLUE \citep{wang2019gluemultitaskbenchmarkanalysis}\footnote{Find a complete overview of the BabyLM evaluation pipeline at \url{https://github.com/babylm/evaluation-pipeline-2025}.}. While these benchmarks provide broad coverage of linguistic competence, they do not directly and comprehensively assess pragmatic abilities. Given the developmental motivation behind BabyLMs, we argue that it is equally important to examine whether such models exhibit the pragmatic reasoning observed in humans. 

\paragraph{Pragmatic Evaluation in LMs.}
As pragmatic knowledge is essential for successful communication, recent studies have explored whether LLMs exhibit pragmatic reasoning. 
Some studies report that LLMs can perform competitively with humans on specific tasks such as metaphor comprehension \citep{hu-etal-2023-fine, sanchez-bayona-agerri-2025-metaphor}, but many find that they still struggle with a wide range of phenomena, including sarcasm and jokes \citep{hu-etal-2023-fine, jentzsch-kersting-2023-chatgpt}, theory of mind \citep{Shapira2023, Trott2023, Gandhi2023}, implicit causality \cite{sieker-etal-2023-beyond, Kankowski2025}, context-dependent reference resolution \citep{junker-etal-2025-multimodal, ma2025visionlanguagemodelspragmaticallycompetent}, or inferences like presuppositions \citep{KabbaraChang2022, sieker-zarriess-2023-antipsps, tsvilodub2024experimentalpragmaticsmachinestesting, Sieker2025, lachenmaier-etal-2025-llms}.

When it comes to the Gricean maxims, \citet{hu-etal-2023-fine}, for example, evaluated LLMs' ability to understand intended meanings by prompting models with short English stories and asking what a character wanted to convey by flouting a maxim, given a set of possible answers. They found that the models would generally assign higher probabilities to literal meanings over the speaker's intended meaning. Similarly, LLMs demonstrated bias towards literal meanings during a pragmatic evaluation for Korean language by \citet{park-etal-2024-pragmatic}. In their experiment, models performed poorly on the maxim of Relation and well on the maxim of Quality when selecting the pragmatic meanings from given options, but showed reversed patterns for open-ended questions about the speaker’s intent. \citet{shisen-etal-2024-large}, on the other hand, evaluated models' ability to infer implicated meanings in multi-turn Chinese dialogues and found no significant variation in model performances across maxims. Moreover, most models failed to generate correct interpretations for implicatures despite being able to identify them in a multiple-choice setting. Other examples of pragmatic evaluations of LLMs by incorporating implicatures include (but are not limited to) \citet{zheng-etal-2021-grice}, who presented the GRICE dataset for assessing the pragmatic reasoning of LLMs while taking into account other aspects of modern dialogue modeling like coreference; \citet{cho-ismkim99-skku-edu-2024-pragmatic}, who compared cosine similarities of literal meanings of scalar implicatures with their pragmatic meanings; and \citet{kurch-etal-2024-large}, who tested whether LLMs can derive atypicality inferences that are triggered through information redundancy.

Building on this line of work, we extend prior studies on LM pragmatic competence and Gricean maxims by introducing a child-directed, maxim-balanced benchmark that enables direct comparison between model and child performance.  Inspired by \citet{article}'s CVT, we compile a dataset of 2,250 conversational items in a controlled forced-choice format. Our benchmark is particularly well-suited to the BabyLM Challenge because its simple, child-appropriate language and controlled design offer a fine-grained, diagnostic test of pragmatic abilities, while minimizing reliance on large-scale training data or extensive world knowledge.

\section{Approach}\label{Approach}

In order for pragmatic interpretations (those that go beyond literal ones) to arise, a listener must know the rules of conversation, recognize when they are violated, and discern when an utterance may be literally unfitting (uncooperative) yet pragmatically acceptable (cooperative). 
Building on the framework of Gricean maxims, \citet{article} examined the pragmatic competence of children by testing whether they could identify an uncooperative (i.e., maxim-violating) answer among a pair of responses to a given question.
We adopt this same forced-choice paradigm to evaluate the pragmatic sensitivity of language models.

\paragraph{Data.}

We base our evaluation on the CVT 
and extend this resource into a large-scale benchmark by generating additional CVT-style items with GPT-4 \citep{openai2024gpt4technicalreport} and manually curating the outputs to maintain child-level vocabulary\footnote{This is derived from the fact that the pretraining data for baseline BabyLMs mostly consists of input received by children.}, grammaticality, naturalness and adherence to the targeted maxim.
The final dataset contains 2,250 dialogues, balanced across five maxims: Quantity I, Quantity II, Quality, Relation, and Politeness.
Full details of the augmentation process and quality control criteria are provided in Appendix \ref{appendix:data_augmentation}. Also, see Appendix \ref{appendix:data_examples} for examples of experimental items of our dataset. 

\paragraph{Models.} We use the following baseline BabyLMs pretrained on BabyLM corpora that were released in two tracks (Strict for models trained on at most 100M tokens and Strict-small for models trained on at most 10M tokens)\footnote{Baseline models for previous years and this year's submission are available at \url{https://huggingface.co/babylm} and \url{https://huggingface.co/BabyLM-community}.}: two auto-regressive LMs, namely  GPT-2 \citep{radford2019language} and Baby Llama \citep{timiryasov2023babyllamaknowledgedistillation} and two masked LMs, namely LTG-BERT \citep{samuel-etal-2023-trained} and Roberta \citep{liu2019robertarobustlyoptimizedbert}. Finally, to assess the effect of more training data on the pragmatic performance of language models and to enable a comparison with an LLM, we evaluate the decoder-only OLMo-1B \citep{groeneveld-etal-2024-olmo} as a representative of fully open LLMs that has been trained on 3T tokens. 

\paragraph{Evaluation.} 

Using our curated dataset, we evaluate pragmatic sensitivity of language models in an unsupervised setting. Specifically, we measure a model’s sentence acceptability for the two candidate answers to a question: one that follows a Gricean maxim (follower) and one that violates it (violator). 
For incremental models, we compute the conditional log-probability of the answer given the question, while in the case of masked language models, we use the improved pseudo-log-likelihood proposed by \citet{kauf2023betterwaymaskedlanguage}. In both cases, the probability of the answer is calculated as the sum of the log-probabilities of its tokens, normalized by its length.
For each item, we assess whether the model assigns a higher probability to the maxim-following answer:\\
\resizebox{\columnwidth}{!}{$
\mathbbm{1}\!\left[ P(\text{Answer}_{\text{Follower}} \mid \text{Question}) > P(\text{Answer}_{\text{Violator}} \mid \text{Question}) \right]
$}\\
Model accuracy for each maxim is defined as the proportion of items for which the model assigns a higher probability to the maxim-follower response. 
We obtain models' scores through Minicons \citep{misra2022miniconsenablingflexiblebehavioral}\footnote{Available at
\url{https://github.com/kanishkamisra/minicons}.}, which is is an open-source library for extracting sentence acceptability measures in language models.

\section{Results}

\definecolor{lightgray}{gray}{0.95}

\setlength{\tabcolsep}{6pt}

\begin{table*}[h!]
    \centering
    \small
    \rowcolors{2}{lightgray}{white}
    \begin{tabularx}{\textwidth}{ll*{6}{>{\centering\arraybackslash}X}}
    \toprule
        & & \textbf{Quantity\,I} & \textbf{Quantity\,II} & \textbf{Quality} & \textbf{Relation} & \textbf{Politeness} & \textbf{Overall} \\ 
    \midrule
        
    \multicolumn{8}{l}{\textbf{Strict-small BabyLMs}} \\
        & GPT-2 & \textbf{0.59} & \textbf{0.59} & 0.61 & \textbf{0.66} & \textbf{0.80} & \textbf{0.65}\\ 
        & Baby Llama & 0.56 & 0.49 & 0.74 & 0.63 & 0.68 & 0.62\\ 
        & LTG BERT & 0.45 & 0.45 & \textbf{0.76} & 0.58 & 0.46 & 0.54\\ 
        & Roberta & 0.33 & 0.36 & 0.64 & 0.60 & 0.57 & 0.50\\ 
        & \textit{Average} & 0.48 & 0.47 & 0.69* & 0.61 & 0.63 & \\ 
    \midrule
        
    \multicolumn{8}{l}{\textbf{Strict BabyLMs}} \\
        & GPT-2 & 0.60 & 0.68 & 0.76 & 0.72 & \textbf{0.76}  & \textbf{0.70}\\ 
        & Baby Llama & 0.55 & \textbf{0.64} & 0.75 & 0.67 & 0.70 & 0.66\\ 
        & LTG BERT & \textbf{0.64} & 0.57 & \textbf{0.79} & \textbf{0.74} & 0.52 & 0.65 \\ 
        & Roberta & 0.51 & 0.47 & 0.67 & 0.68 & 0.59 & 0.58\\ 
        & \textit{Average} & 0.58 & 0.59 &  0.74* & 0.70 & 0.64 & \\ 
    \midrule
        
    \multicolumn{8}{l}{\textbf{LLM}} \\
        & OLMo-1B & 0.76 & 0.83 & 0.83 & 0.84 & 0.67 & 0.79\\ 
    \midrule
        
    \multicolumn{8}{l}{\textbf{Children}} \\
        &  & 0.58 & 0.78 & 1.0 & 1.0 & 0.93 & 0.86\\ 
    \bottomrule
    \end{tabularx}
    \caption{Accuracy scores across the Gricean maxims. \textit{Strict-small} models are pretrained on <10M tokens and \textit{Strict} models on <100M. OLMo-1B is pretrained on 3T tokens. All models are evaluated on 2,250 items, while child accuracy scores are from 8 neurotypical children from \citet{article}. For the Strict-small and Strict groups, the highest score of each maxim is bolded. The highest average across all maxims is marked with (*).}
    \label{tab:results}
\end{table*}

In Table \ref{tab:results}, we report accuracy per maxim for the BabyLM baselines in the Strict-small and Strict tracks. Furthermore, we present the results from OLMo-1B. Finally, as a reference point, we include the results of children who were tested on the CVT by \citet{article}. 
In the following, we break down the results by conversational maxim, model architecture and model size. 

\paragraph{Results by Gricean Maxim.}

As shown in Table \ref{tab:results}, model performance varies considerably across different maxims in both the Strict-small and Strict tracks. 
The maxims Quantity I (\textit{Be informative}) and Quantity II (\textit{Avoid redundant information}) are consistently the most challenging, with average accuracies for all BabyLMs peaking at only 0.59. 
In contrast, Quality (\textit{Be truthful}) emerges as the easiest category for most BabyLMs with the average accuracy as high as 0.74.
Relation (\textit{Be relevant}) and Politeness (\textit{Be polite}) generally fall in between these extremes (although exceptions apply), with the average accuracies above chance but below the best Quality results.

This pattern suggests that factuality is easier for BabyLMs to capture from limited data, likely because it can be learned from explicit statements and lexical associations in the training data, whereas judgments of informativeness and redundancy require more context-sensitive reasoning. 
The representative examples from the original CVT in Table \ref{tab:cvt_examples} illustrate this: in Quantity I, while \textit{"with milk"} falls in the range of pragmatically accepted answers, \textit{"in a cup"} might contain tokens (or token combinations) that are more frequent in the training data. 
Across the other maxims, such frequent continuations may similarly override pragmatic appropriateness in model predictions. 

To quantify how consistently models agree on these difficulty patterns, we ranked maxims per model and computed Kendall’s $W$.
Agreement was high in the Strict-small track ($W = 0.80$) and moderate in the Strict track ($W = 0.68$), indicating that models trained on less data tend to exhibit more similar difficulty patterns. 
In both tracks, the maxims Quantity I and Quantity II were generally ranked as the hardest maxims. Full rankings and statistics are reported in Table \ref{tab:kendall_ranks}.


\begin{table}[t]
\centering
\small
\setlength{\tabcolsep}{4pt}
\begin{tabular}{lccccc}
\toprule
\textbf{Model} & \textbf{Quant.~I} & \textbf{Quant.~II} & \textbf{Qual.} & \textbf{Rel.} & \textbf{Polite} \\
\midrule
\multicolumn{6}{l}{\textit{Strict-small}} \\
GPT-2       & {\textcolor{comp4}{\scalebox{1.5}{$\bullet$}}} & {\textcolor{comp5}{\scalebox{1.5}{$\bullet$}}} & {\textcolor{comp3}{\scalebox{1.5}{$\bullet$}}} & {\textcolor{comp2}{\scalebox{1.5}{$\bullet$}}}  & {\textcolor{comp1}{\scalebox{1.5}{$\bullet$}}}  \\
Baby Llama  & {\textcolor{comp4}{\scalebox{1.5}{$\bullet$}}} & {\textcolor{comp5}{\scalebox{1.5}{$\bullet$}}} & {\textcolor{comp1}{\scalebox{1.5}{$\bullet$}}} & {\textcolor{comp3}{\scalebox{1.5}{$\bullet$}}}  & {\textcolor{comp2}{\scalebox{1.5}{$\bullet$}}}  \\
LTG BERT    & {\textcolor{comp4}{\scalebox{1.5}{$\bullet$}}} & {\textcolor{comp5}{\scalebox{1.5}{$\bullet$}}} & {\textcolor{comp1}{\scalebox{1.5}{$\bullet$}}} & {\textcolor{comp2}{\scalebox{1.5}{$\bullet$}}}  & {\textcolor{comp3}{\scalebox{1.5}{$\bullet$}}}  \\
RoBERTa     & {\textcolor{comp5}{\scalebox{1.5}{$\bullet$}}} & {\textcolor{comp4}{\scalebox{1.5}{$\bullet$}}}& {\textcolor{comp1}{\scalebox{1.5}{$\bullet$}}} & {\textcolor{comp2}{\scalebox{1.5}{$\bullet$}}}  & {\textcolor{comp3}{\scalebox{1.5}{$\bullet$}}}  \\
\midrule
\multicolumn{6}{l}{\textit{Strict}} \\
GPT-2       & {\textcolor{comp5}{\scalebox{1.5}{$\bullet$}}} & {\textcolor{comp4}{\scalebox{1.5}{$\bullet$}}} & {\textcolor{comp1}{\scalebox{1.5}{$\bullet$}}} & {\textcolor{comp3}{\scalebox{1.5}{$\bullet$}}}  & {\textcolor{comp2}{\scalebox{1.5}{$\bullet$}}}  \\
Baby Llama  & {\textcolor{comp5}{\scalebox{1.5}{$\bullet$}}} & {\textcolor{comp4}{\scalebox{1.5}{$\bullet$}}} & {\textcolor{comp1}{\scalebox{1.5}{$\bullet$}}} & {\textcolor{comp3}{\scalebox{1.5}{$\bullet$}}}  & {\textcolor{comp2}{\scalebox{1.5}{$\bullet$}}} \\
LTG BERT    & {\textcolor{comp3}{\scalebox{1.5}{$\bullet$}}}  & {\textcolor{comp4}{\scalebox{1.5}{$\bullet$}}} & {\textcolor{comp1}{\scalebox{1.5}{$\bullet$}}} & {\textcolor{comp2}{\scalebox{1.5}{$\bullet$}}}  & {\textcolor{comp5}{\scalebox{1.5}{$\bullet$}}} \\
RoBERTa    & {\textcolor{comp4}{\scalebox{1.5}{$\bullet$}}} & {\textcolor{comp5}{\scalebox{1.5}{$\bullet$}}} & {\textcolor{comp3}{\scalebox{1.5}{$\bullet$}}}  & {\textcolor{comp1}{\scalebox{1.5}{$\bullet$}}} & {\textcolor{comp2}{\scalebox{1.5}{$\bullet$}}}  \\
\bottomrule
\end{tabular}
\caption{
Maxim difficulty rankings for each model ({\textcolor{comp1}{\scalebox{1.5}{$\bullet$}}} = easiest, {\textcolor{comp5}{\scalebox{1.5}{$\bullet$}}} = hardest). 
Kendall’s $W$: Strict-small = 0.80, $\chi^2$(4) = 12.80, $p = 0.012$; 
Strict = 0.68, $\chi^2$(4) = 10.80, $p = 0.029$.}
\label{tab:kendall_ranks}
\end{table}

\begin{figure}[t]
\centering
\includegraphics[width=\linewidth]{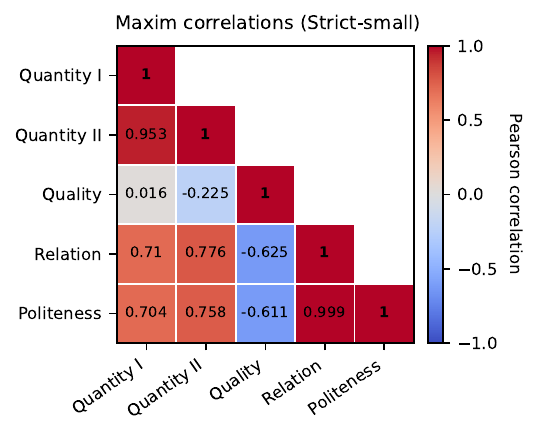}\\[-0.3em]
\includegraphics[width=\linewidth]{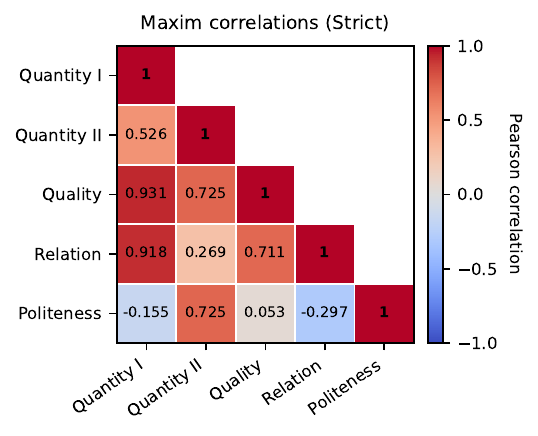}
\caption{Pearson correlations between per-model accuracies across maxims. Top: Strict-small track. Bottom: Strict track.}
\label{fig:maxim_corr_heatmaps}
\end{figure}

\paragraph{Inter-maxim Correlations.}
To examine whether performance on different maxims co-varies across models,
we computed Pearson correlations between per-model accuracies across maxims for each track (Figure \ref{fig:maxim_corr_heatmaps}).
In the Strict-small track, correlations between maxims tend to be extreme. 
Quantity I and Quantity II show an almost perfect correlation ($r = 0.953$). 
The same track also reveals a striking near-perfect correlation between Relation and Politeness ($r = 0.999$), suggesting that topicality and politeness violations may be treated in similar ways. 
In contrast, Quality stands out in the Strict-small track as largely decoupled from the other maxims (near-zero or negative correlations), which may indicate that detecting literal implausibility (e.g., \textit{"on the moon"}, Table \ref{tab:cvt_examples}) behaves independently of other pragmatic abilities under data constraints.
In the Strict track, correlations are overall weaker and more varied, with some negative associations appearing for pairs that were strongly positive in the Strict-small track (e.g., Politeness vs. Quantity I, ($r = -0.155$), possibly reflecting a partial decoupling of politeness from topicality and informativeness once models have more data. Notably, the maxim of Quality strongly correlates with Quantity I in the Strict track ($r = 0.931$), despite being among the easiest maxims in difficulty rankings (Table \ref{tab:kendall_ranks}), showing that correlation patterns capture co-variation rather than absolute difficulty.
Overall, the shift from rather extreme relations in the Strict-small track to more varied and generally weaker associations in the Strict track complements the prior results, suggesting that with more training data, models begin to differentiate more between pragmatic dimensions.\\


\paragraph{Effects of Architecture and Model size.}
Referring to Table \ref{tab:results}, it is further notable that across both tracks, GPT-2 consistently outperforms LTG-BERT and RoBERTa models, with BabyLLaMA typically ranking second. RoBERTa BabyLM performs worst overall, particularly on the Quantity maxims. 
Increasing the training data from the Strict-small (<10M tokens) to the Strict (<100M tokens) track generally improves performance, especially for Quantity I and Quantity II, where average scores increase by 0.09–0.12. Gains for Quality and Relation are more modest, while Politeness scores remain similar across tracks. 
Interestingly, GPT-2’s Politeness score decreases in the Strict track, suggesting that greater exposure to varied language might introduce alternative patterns that increase uncertainty in politeness judgments. 

The large-scale OLMo 1B model substantially outperforms all BabyLM baselines across the maxims, with the exception of GPT-2 Strict-small, which scored 0.80 in Politeness. OLMo 1B scored substantially higher on the maxim of Relation compared to BabyLMs, indicating that sensitivity to topic relevance tends to emerge with increased training data.

The differences of scores may also reflect how auto-regressive and masked LMs handle conversational flow: in our setup, probability assignment to an entire answer benefits from modeling sequences as coherent continuations rather than token-masked completions. The performance gap between Strict-small and Strict models also indicates that increased training data helps, but does not eliminate the persistent difficulties with Quantity-related judgments.
    

\paragraph {Comparison to Child Performance.}
In \citet{article}, neurotypical children showed an overall high accuracy (0.86), likely reflecting the development of Theory of Mind \citep{BARONCOHEN198537}, the impairment of which damages recognizing speaker’s intended meaning. 
BabyLMs share some similarities with children: like them, they perform best on the maxim of Quality and worst on the Quantity maxims (Table \ref{tab:results}).
However, they do not demonstrate the the same high performance as children in maxims of Relation and Politeness.
Furthermore, children’s overall high accuracy indicates that, by school age, they are already highly sensitive to conversational norms. In contrast, BabyLMs reach only 0.50–0.70 of overall accuracy, underscoring a substantial gap in pragmatic competence between small-scale LMs and six-to-seven-year-old human speakers.
The LLM shows a different pattern: it surpasses children on the Quantity maxims, indicating stronger performance on informativeness-related judgments, but still falls short on other maxims, especially Politeness, suggesting that socially grounded pragmatic norms do not emerge automatically from large-scale pretraining.
Overall, both small-scale and large-scale models reveal persistent limitations in capturing the full range of conversational norms.


\paragraph{Summary and Discussion.}
Across both data tracks, the greatest deficits in BabyLM performance concerned judgments of informativeness required for an appropriate response. 
The maxim of Quality was the easiest, with Relation and Politeness in between (Tables \ref{tab:results} and \ref{tab:kendall_ranks}). 
Correlation analyses further revealed that, under severe data constraints, most maxims were strongly associated with at least one other (e.g., Quantity I and II; Relation and Politeness) (Figure \ref{fig:maxim_corr_heatmaps}). These associations tended to weaken with more data, indicating a shift toward more differentiated treatment of pragmatic categories.
Among architectures, GPT-2 performed best overall, RoBERTa worst, and scaling from <10M to <100M tokens yielded the largest gains on Quantity, though sometimes at the expense of Politeness (Table \ref{tab:results}), suggesting that autoregressive modeling and modest scaling benefit informativeness but may reduce social-pragmatic sensitivity.
Compared to children, (0.86 overall, Table \ref{tab:results}), BabyLMs mirrored the relative ordering of difficulty but 
scored substantially lower (0.50–0.70). 
The LLM (OLMo-1B) outperformed BabyLMs' overall performance in all maxims and exceeded child performance on Quantity, yet lagged considerably on the remaining maxims, showing that large-scale pretraining enhances information-structuring abilities but offers limited gains in other dimensions of pragmatic understanding.

Our results align with earlier findings that pragmatic competence in language models scales with model size and training data but may remain below human levels. For instance, \citet{hu-etal-2023-fine} reported that GPT-2 with 117M parameters did not perform above chance when interpreting maxim-flouting utterances, in line with our observation that BabyLMs trained on <100M tokens perform rather poorly across maxims.
At the other end of the scale, \citet{shisen-etal-2024-large} found that LLaMA 2 models with 13B parameters performed above chance across several Gricean maxims but still achieved only about half the human score, while GPT-4 matched human performance. 
In this context, our results with OLMo-1B suggest that large-scale pretraining can surpass child performance on informativeness but still leaves substantial gaps on more socially grounded maxims such as Politeness. 

Overall, these findings indicate that scaling data and parameters improves some aspects of pragmatic reasoning in language models, while their absolute performance remains far from child-like.  
This underscores the importance of dedicated evaluation benchmarks targeting pragmatic abilities, ensuring that the developmental goals of the BabyLM challenge address this crucial aspect of language use.


\section{Conclusion}
This paper introduced a novel large-scale benchmark for evaluating the pragmatic competence of language models, grounded in the Gricean maxims and adapted from a psycholinguistic test suite.
Using 2,250 conversational items, we assessed BabyLM baseline models trained on constrained data alongside a large-scale 3T-token model and projected their performance on that of children.
Our results indicate that increasing training data from <10M to <100M tokens leads to performance gains, yet BabyLMs remain below child-level competence. They demonstrated the lowest performance in assessing the appropriate amount of information for conversationally acceptable responses and did not exhibit a preference for answers that were more polite or contextually relevant. Furthermore, correlation analyses revealed that under data-limited conditions, models tend to conflate certain pragmatic competences, but these associations weaken with more data, suggesting that additional exposure allows models to more clearly differentiate between distinct pragmatic dimensions.

Our benchmark provides a linguistically grounded, scalable evaluation resource that enables systematic and comparable measurement of pragmatic behavior across models of different sizes and training regimes. By extending the BabyLM evaluation suite with a dedicated pragmatic benchmark, this work provides a tool for systematically tracking progress on this essential aspect of human-like language understanding.

\paragraph{Limitations and Future Directions.}

In this section, we state the limitations of our study and possible directions they offer for future work.

First, unlike other datasets \citep{zheng-etal-2021-grice,hu-etal-2023-fine, park-etal-2024-pragmatic}, the conversational items in our dataset do not include detailed scenarios that are embedded before prompting models with dialogues. In certain contexts, the maxim-violator responses in our dataset could be in fact appropriate; for instance, the answer \textit{"My best friend is Peter. He wears clothes."} (Table \ref{tab:cvt_examples}) would not be redundant in a scenario where others are unclothed.
However, even within a minimal-context setting like the one implemented in this study, non-linguist participants have been shown to consistently favor responses that adhere to conversational maxims. For example, \citet{okanda2015understanding} applied a revised Japanese version of the CVT and found that adults were able to identify non-cooperative answers and articulate the reasoning behind their judgments. Nevertheless, future work could expand our conversational items by introducing explicit context that would render maxim-violator responses cooperative (as in the above-mentioned example) to examine whether such framing would change model preferences.

Second, we acknowledge that model probability assignments may be influenced by the distributional properties of tokens independent of context, which can make evaluations based on sequence scores prone to bias. Future work could address this by expanding the dataset to include a wider range of lexical variations.
\newline Third, we selected the 1B-parameter OLMo model due to computational constraints. Although our primary focus was on BabyLMs rather than large models, evaluating systems of varying sizes offers valuable comparative insights. Furthermore, as our results suggest that model architecture affects pragmatic performance, future work could test whether these patterns hold for other architectures, such as instruction-tuned or multimodal (Baby) models. 

Finally, our dataset is currently limited to English; therefore, extending this evaluation to multilingual settings would allow for more robust conclusions and enable meaningful cross-linguistic comparisons of pragmatic competence.

\section*{Ethical Statement}
Our work uses publicly available data from a psycholinguistic study on children; we do not conduct any new experiments involving human subjects. No personal information from the original study is shared, except for the mean age of participants as reported by the authors. We do not train any new models; instead, our evaluation dataset was automatically generated and subsequently reviewed and refined by two of the authors. The dataset consists of short, child-level conversational exchanges, and we believe that none of the items raise ethical concerns or reinforce biases toward sensitive groups. Our evaluation focuses exclusively on the pragmatic competence of language models. We do not address other potential harms or limitations of these systems such as discrimination, toxicity and misinformation \citet{weidinger2021ethical} which remain important areas for continued investigation and responsible deployment. We share our evaluation dataset, code and model results publicly to facilitate future use and promote transparency.

\section*{Acknowledgements}
We warmly thank the anonymous reviewers for their insightful comments. We acknowledge support from several projects and funding institutions: 1) “SAIL: SustAInable Life-cycle of Intelligent Socio-Technical Systems" (Grant ID NW21-059A), an initiative of the Ministry of Culture and Science of the State of Northrhine Westphalia; 2) Erasmus+, a European Union funding that facilitates periods of study or training abroad, 3) Deutsche
Forschungsgemeinschaft (DFG, German Research Foundation) – CRC-1646, project number
512393437, project B02.

\appendix

\newpage
\section{Appendix}\label{sec:appendix}

\subsection{Dataset Augmentation Details} \label{appendix:data_augmentation}

In order to expand CVT for model evaluation, we employed the GPT-4 chat interface \citep{openai2024gpt4technicalreport} to synthesize similar items.
First, we provided GPT-4 with a brief description of the maxims and examples from the CVT, and asked it to generate 25 corresponding dialogue templates, each accompanied by two paraphrased versions of the question. Answers were unchanged in the paraphrased versions; this was due to some answers being too short to paraphrase, and we chose a unified method for all items.
Next, each template was presented with its original CVT example, and GPT-4 was asked to produce four new conversations per template, using vocabulary appropriate for children. Finally, the paraphrased versions of both the original CVT items and the newly created conversations were generated based on their respective templates. This resulted in a dataset consisting of 25 CVT items and 350 GPT-generated ones. Table \ref{table:1} shows a concrete example of this pipeline. 

\begin{table*}[h]
\small
\begin{tabularx}{0.9\textwidth} { 
   >{\raggedright\arraybackslash}p{4.5cm} 
   >{\raggedright\arraybackslash}X 
   >{\raggedright\arraybackslash\hspace{0pt}}p{4.5cm}}
\toprule
 \textbf{Stage} & \textbf{Example} \\
 \midrule
 \textbf{1. GPT-4 is prompted with a} & Question: \textit{What would you like to buy in this shoe-shop?}\\
\textbf{dialogue from CVT} & Follower: \textit{A pair of trainers.}\\
 & Violator: \textit{A pair of shoes.}\\
 \midrule
 \textbf{2. GPT-4 generates a template}  & Question: \textit{What would you like to buy in this} \texttt{[type of store]}?\\
\textbf{from the dialogue} & Follower (specific, informative): \texttt{[A concrete item you might buy in that specific  store]}\\
 & Violator (vague, unhelpful): \texttt{[A general category that lacks specific information]}\\
 \midrule
\textbf{3. GPT-4 generates four new} & 1. Question: \textit{What would you like to buy in this toy shop?}\\
\textbf{dialogues from the template of} & Follower: \textit{A teddy bear.}\\
\textbf{ stage 2} & Violator: \textit{A toy.}\\ 
 & 2. Question: \textit{What would you like to buy in this bakery?}\\
 & Follower: \textit{A chocolate muffin.}\\
 & Violator: \textit{Something to eat.}\\ 
 & 3. Question: \textit{What would you like to buy in this bookshop?}\\
 & Follower: \textit{A Harry Potter book.}\\
 & Violator: \textit{A book.}\\
 & 4. Question: \textit{ What would you like to buy in this clothes shop?}\\
 & Follower: \textit{A red jacket.}\\
 & Violator: \textit{Some clothes.}\\
 \midrule
\textbf{4. GPT-4 generates 2} & 1. \textit{Is there something you'd like to get from this} \texttt{[type of store]}? \\
\textbf{paraphrased questions of the} & 2. \textit{What are you looking for in this} [type of store]?\\
\textbf{template of stage 2} & \\
 \bottomrule
\textbf{5. GPT-4 generates 2} & 1. Question: \textit{Is there something you'd like to get from this shoe-shop?}\textcolor{white}{?}\\
\textbf{paraphrased versions of the} & Follower: \textit{A pair of trainers.}\\
\textbf{original CVT dialogue given in} & Violator: \textit{A pair of shoes.}\\
\textbf{stage 1 and all other dialogues} & 2. Question: \textit{What are you looking for in this shoe-shop?}\textcolor{white}{?}\\
\textbf{generated in stage 3} & Follower: \textit{A pair of trainers.}\\
 & Violator: \textit{A pair of shoes.}\\
 & And so on\\
\end{tabularx}
\caption{An example of the data augmentation pipeline for the maxim of Quantity I (\textit{Be informative}). The Follower
adheres to the maxim, while the Violator does not.
For each dialogue in CVT, four more examples with child-level vocabulary were created. Later, two paraphrased versions for all dialogues (those from CVT and GPT-generated ones) were synthesized and added.}  
\label{table:1}
\end{table*}

Two of the authors manually reviewed all 375 items and made adjustments based on the following criteria: 
\begin{itemize}
    \itemsep-0.1cm 
    \item The follower’s answer does not follow the maxim.
    \item The violator’s answer does not violate the maxim.
    \item The paraphrased versions do not correctly paraphrase the questions.
    \item The words exceed child-level vocabulary.
    \item The questions and/or answers are unnatural or ungrammatical.
    \item The answers are inadequate for model evaluation.
\end{itemize}

The last criterion reflects our effort to minimize superficial differences between the follower and violator answers wherever possible. 
Since our evaluation compares model probabilities for each answer pair, irrelevant lexical differences would distort the results.
For example, in response to the question \textit{Who is your best friend?}, if one answer was \textit{My best friend is \textbf{John}. He goes to my school.} and the other was \textit{My best friend is \textbf{Peter}. He wears clothes.}, not replacing the names with a single name would introduce noise unrelated to pragmatic reasoning.
However, in some cases, such standardizations were not applicable due to the nature of the maxims being tested; for example, in the case of the maxim of Relation, altering the content is necessary for the violation.

Once approved, we diversified the dataset by assigning speaker names (Leslie, Joan, and Thomas) and rotating them such that each conversation had one name as the questioner and another as the responder, used consistently across both responses. The names were randomly selected as English common names from the firstname database at \url{https://github.com/KarlAmort/firstname-database}.
This process yielded a final dataset of 2,250 conversational items, since six variations with names were generated from each of the 375 curated conversations. 

\newpage \subsection{Dataset Examples} \label{appendix:data_examples}

We depict a few examples from the dataset. The maxim-violator's answer is marked with (*).\newline

\textbf{Maxim of Quantity I (Be informative):}
\begin{itemize}
    \item Leslie: What did you eat for supper?\\Thomas: Tomato soup.\\Thomas: A dish.* 
    \item Leslie: What did you see at the zoo?\\Joan: The lions.\\Joan: Some animals.* 
    \item Leslie: What did you get for Christmas?\\Joan: A gift.*\\Joan: A toy train. 
    \item Joan: How do you prefer your pancakes?\\Leslie: On a plate.*\\Leslie: With maple syrup.
\end{itemize}

\textbf{Maxim of Quantity II (Avoid redundant information):}
\begin{itemize}
    \item Leslie: Who is your neighbor?\\Thomas: My neighbor is Mr. Tom. He has a dog.\\Thomas: My neighbor is Mr. Tom. He lives in a house.*
    \item Joan: What pet do you like?\\Leslie: I like puppies and kittens which are pets.*\\Leslie: I like puppies and kittens which are cute. 
    \item Leslie: Where did you go last weekend?\\Joan: I went to grandma’s house and baked cookies.\\Joan: I went to grandma’s house and I didn’t stay in my room.* 
    \item Joan: Which is your favourite fruit?\\Thomas: Watermelon which is a fruit.*\\Thomas: Watermelon which is juicy.
\end{itemize}

\textbf{Maxim of Quality (Be truthful):}
\begin{itemize}
    \item Thomas: Is there any more popcorn?\\Leslie: Yes, there’s a bowl in the kitchen.\\Leslie: Yes, it’s raining popcorn outside.*
    \item Leslie: Where do you do your homework?\\Joan: I do them on a dragon’s back.*\\Joan: I do them in my room.   
    \item Joan: Do you have any pets?\\Leslie: Yes, I have a cat and a fish.\\Leslie: Yes, I have a thousand elephants.*   
    \item Leslie: Why don’t you come outside?\\Thomas: Because I’m helping mom bake.\\Thomas: Because I’m on a spaceship.*
\end{itemize}

\textbf{Maxim of Relation (Be relevant):}
\begin{itemize}
    \item Joan: What did you do on the weekend?\\Thomas: I went to the zoo.\\Thomas: My socks are green.*
    \item Thomas: What do you like to play?\\Leslie: I like to play tag.\\Leslie: I like chocolate cake.*
    \item Joan: What is your favourite animal?\\Thomas: I like pencils best.*\\Thomas: I like pandas best.
    \item Thomas: What songs do you know?\\Joan: I know “Twinkle Twinkle Little Star.\\Joan: I know how to tie my shoes.*
\end{itemize}

\textbf{Maxim of Politeness (Be polite):}
\begin{itemize}
    \item Thomas: Do you like my new haircut?\\Joan: It looks awful.*\\Joan: It looks nice.
    \item Thomas: Would you like to try some of my cake?\\Joan: No, thanks.\\Joan: No, it’s disgusting.*
    \item Joan: May I use your calculator?\\Thomas: No, don’t touch my stuff.*\\Thomas: No, sorry, I need it right now, but you can use it after.
    \item Leslie: Could you help me with my puzzle?\\Thomas: Do it by yourself.*\\Thomas: Sure, after I finish this one.
\end{itemize}


\end{document}